\documentclass[a4paper,twoside]{article}
\pdfoutput=1

\usepackage{epsfig}
\usepackage{subcaption}
\usepackage{calc}
\usepackage{amssymb}
\usepackage{amstext}
\usepackage{amsmath}
\usepackage{amsthm}
\usepackage{multicol}
\usepackage{pslatex}
\usepackage{apalike}
\usepackage{algorithm2e}
\usepackage[bottom]{footmisc}

\usepackage{multirow}
\usepackage{breqn}
\usepackage{tikz}
\usetikzlibrary{spy}
\usetikzlibrary{calc}
\usetikzlibrary{positioning}

\usepackage{SCITEPRESS} 

\begin{document}

\title{Towards Better Morphed Face Images without Ghosting Artifacts}

\author{\authorname{Clemens Seibold\sup{1}\orcidAuthor{0000-0002-9318-5934}, 
Anna Hilsmann\sup{1}\orcidAuthor{0000-0002-2086-0951} and 
Peter Eisert\sup{1,2}\orcidAuthor{0000-0001-8378-4805}}
\affiliation{\sup{1}Fraunhofer HHI, Berlin, Germany}
\affiliation{\sup{2}Humboldt University of Berlin, Berlin, Germany}
\email{\{clemens.seibold, anna.hilsmannn, peter.eisert\}@hhi.fraunhofer.de}
}

\keywords{Face Morphing Attacks, Ghosting Artifact Prevention, Morphed Face Images Dataset}

\abstract{Automatic generation of morphed face images often produces ghosting artifacts due to poorly aligned structures in the input images. Manual processing can mitigate these artifacts. However, this is not feasible for the generation of large datasets, which are required for training and evaluating robust morphing attack detectors. In this paper, we propose a method for automatic prevention of ghosting artifacts based on a pixel-wise alignment during morph generation. We evaluate our proposed method on state-of-the-art detectors and show that our morphs are harder to detect, particularly, when combined with style-transfer-based improvement of low-level image characteristics. Furthermore, we show that our approach does not impair the biometric quality, which is essential for high quality morphs.}

\onecolumn \maketitle \normalsize \setcounter{footnote}{0} \vfill

\section{INTRODUCTION}
A morphed face image is a composite image that is generated by blending facial images of different subjects.
Since the feasibility of tricking a facial recognition system to match two random subjects with one morphed face image was demonstrated by \cite{Ferrara14}, a significant amount of research has been conducted in generating and detecting such images. Early publications on Morphing Attack Detection (MAD) relied on manually generated morphed face images for training and evaluation. However, since manual generation is a time-consuming task, automatic approaches paved the way for developing data-demanding machine learning-based detectors and evaluations on large datasets.
 
\begin{figure*}[!htb]
	\center
 \begin{subfigure}[b]{0.22\textwidth}
    \begin{tikzpicture}[spy using outlines={rectangle}, node distance=2mm and 0mm, inner sep=0mm]
		\node (I) {\includegraphics[width=1.0\textwidth]{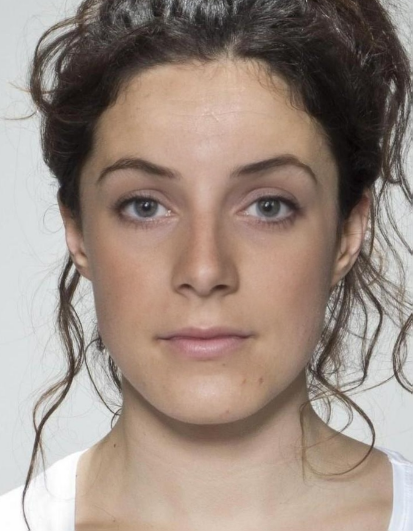}};
        \spy[red,magnification=4,size=0.6\textwidth]  on (0.55, 0.55) in node [below= of I](spy1);
	\end{tikzpicture}		
    \subcaption{Simple Morph}
 \end{subfigure}
 \begin{subfigure}[b]{0.22\textwidth}
	\begin{tikzpicture}[spy using outlines={rectangle}, node distance=2mm and 0mm, inner sep=0mm]
		\node (I) {\includegraphics[width=1.0\textwidth]{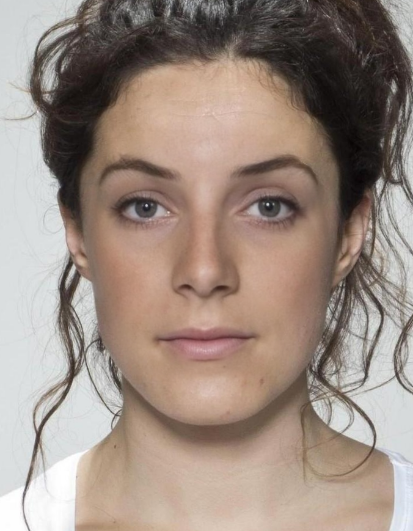}};
		\spy[red,magnification=4,size=0.6\textwidth]  on (0.55, 0.55) in node [below= of I](spy1);
	\end{tikzpicture}
 \subcaption{Proposed Improvement}
 \end{subfigure}
 \begin{subfigure}[b]{0.22\textwidth}
	\begin{tikzpicture}[spy using outlines={rectangle}, node distance=2mm and 0mm, inner sep=0mm]
	\node (I) {\includegraphics[width=1.0\textwidth]{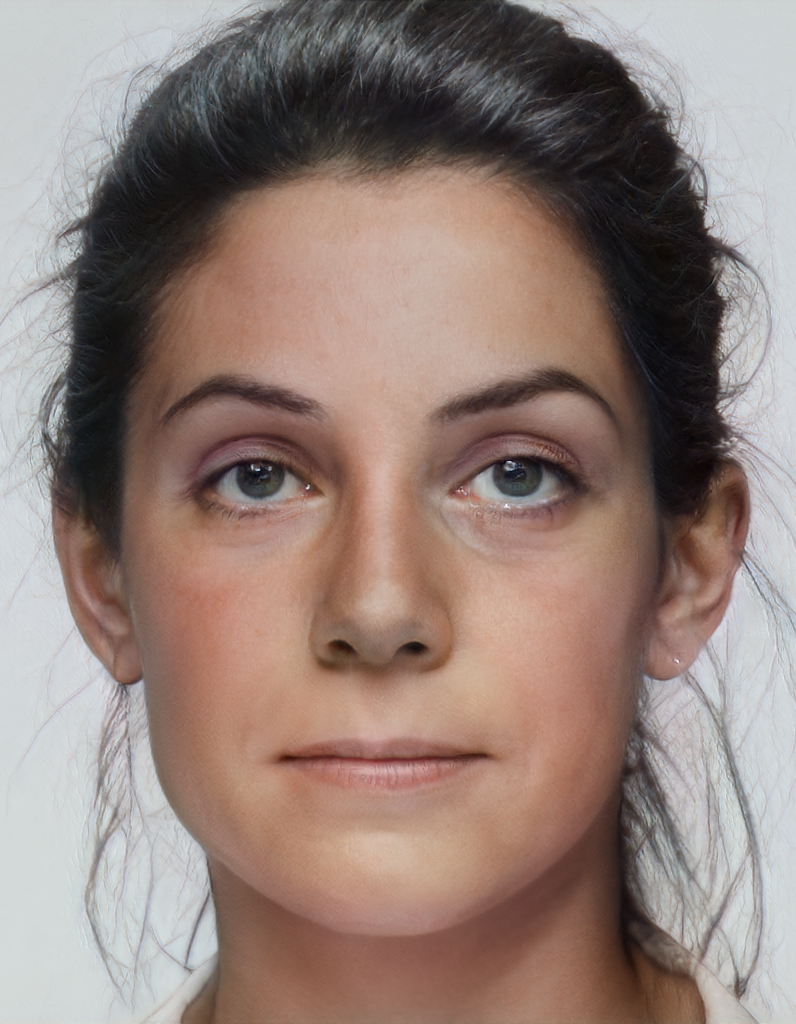}};
	\spy[red,magnification=4,size=0.6\textwidth]  on (0.55, 0.20) in node [below= of I](spy1);
	\end{tikzpicture}
    \subcaption{GAN-based Morph}
 \end{subfigure}
 \begin{subfigure}[b]{0.22\textwidth}
    \begin{tikzpicture}[spy using outlines={rectangle}, node distance=2mm and 0mm, inner sep=0mm]
		\node (I) {\includegraphics[width=0.75\textwidth]{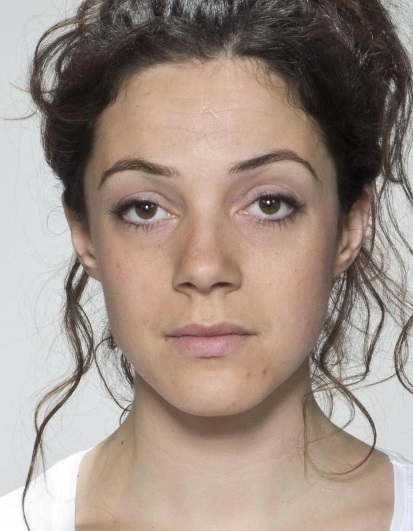}};
        \node[below= of I] (I2) {\includegraphics[width=0.75\textwidth]{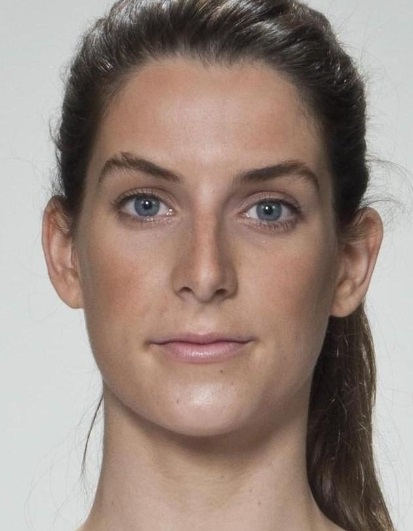}};
	\end{tikzpicture}		
    \subcaption{Input Images}
 \end{subfigure}
 \\
	\caption{Morphed face image generated with a simple keypoint-based method (a), improved with the proposed ghosting artifact prevention method (b), generated with a GAN-based method (c), and input images (d). The ghosting artifact only appears in the simple approach. The GAN-based morph suffers from different artifacts, e.g., the unusual iris and pupil shapes.}
	\label{fig:exampleMorphs}
\end{figure*}

Most automatic face morphing approaches estimate the positions of facial landmarks in both input images~\cite{Makrushin17}, warp the images such that the landmarks have the same shape and position and then additively blend them. Images generated by these methods often suffer from ghosting artifacts caused by inaccuracies in the landmark position estimation or unalignable facial structures. 
These artifacts occur when two structures, such as the iris border, are not perfectly aligned.
Figure \ref{fig:exampleMorphs} shows an example of a ghosting artifact. In the simple morphed face image, a second translucent border of the iris is visible due to inaccurate alignment of the iris shape. Our proposed method, however, prevents the appearance of such artifacts.

An alternative to the key-point-based method is the use of Generative Adversarial Networks (GANs) \cite{Zhang21}. Images generated using GANs do not contain ghosting artifacts, but come with other limitations. For example, the resolution of these images is determined by the GAN architecture, the generation process is hard to control, and they often leave GAN-typical artifacts that allow their detection~\cite{Zhang19}.

In this paper, we address the prevention of ghosting artifacts in automatic key-point-based generation of morphed face images. In real attacks, attackers may manually correct the images to avoid ghosting artifacts, but this approach is impractical for large datasets. Unlike GAN-based methods, our proposed method allows results of any resolution. It uses a pixel-wise alignment technique that maps similar structures, such as the contour of the nostrils, iris, specular highlights etc. such that they have the same shape and position in both input images. Thus, it prevents ghosting artifacts in the final morphed face image. Figure \ref{fig:exampleMorphs} shows an example of a morphed face image generated using a simple key-point-based approach, our proposed improvement method, and a GAN-based approach. 

We evaluate the impact of this enhancement on the detection rate of different state-of-the-art single image-based MAD approaches using uncompressed and compressed images, since studies showed that compressed images are harder to detect and differ strongly in feature space~\cite{Seibold19_WIFS}. Furthermore, we analyze its effect on the objective of face morphing attacks: Creating a face image that looks similar to two different subjects.

To evaluate our approach, we address the following four research questions:\\
\textbf{R1:} Does our proposed method make morphs harder to detect for single-image MAD techniques?\\
\textbf{R2:} Can detectors adapt to these novel morphed face images?\\
\textbf{R3:} Do the improved morphed face images still impersonate two different subjects?\\
\textbf{R4:} Does our proposed method still affect the detection rate of MAD techniques in compressed images?\\

In summary, our contributions are:
\begin{itemize}
	\item A method to prevent ghosting artifacts in morphed face images as an additional component for key-point-based morphing pipelines.
	\item A novel dataset of faultless morphed face images\footnote{Accessible under https://cvg.hhi.fraunhofer.de}.
	\item An evaluation of different morphing methods on state-of-the-art detectors.
\end{itemize}

The paper is structured as follows. The next section describes our pixel-wise improvement method. The experiments, including a short description of the used detectors and datasets, are presented in Section 3. Section 4 provides results of selected MAD techniques and the evaluation of the biometric quality of the used morphed face images. 

\section{RELATED WORK}
Early research on morphing attack generation to study the feasibility of this attack and the detection of such images relied on manually generated morphed face images~\cite{Ferrara14,Ramachandra16}. 
\cite{Makrushin17} proposed an automatic face morphing pipeline to generate visually faultless morphed face images, pushing the automatic generation of large data sets of morphed face images for the development of data-driven detection methods and their evaluation on large datasets. 
Several researchers adopted this concept and trained and evaluated their morphing detection methods on automatically generated morphed face images. However, only a very few authors have published their morphed face images or code for the generation of such. See \cite{Hamza22} for an overview of the generation and detection of morphed face images. 

The mandatory blending process in face morphing pipelines usually impairs the quality of the images, often dampening the high spatial-frequency details, and causing the blended images to appear more dull than the input images. \cite{Seibold21} proposed a method based on style transfer to counter this effect and showed that their improved attacks are harder to detect. 

Other approaches to generate morphed face images are based on GANs. The first approaches were only capable of generating images in small resolutions, such as $64\times64$ pixels \cite{Damer18}. Later approaches benefited from advances in GAN-based image generation, and \cite{Zhang21} proposed a face morphing method based on Style-GAN2~\cite{Karras20}, which can create realistic face images in a resolution of $1024\times 1024$ pixels.

\section{PIXEL-WISE ALIGNMENT FOR MORPHED FACE IMAGE GENERATION}

A typical face morphing pipeline consists of three main components: key-point-based alignment, additive blending, and an optional post-processing step to handle the background. We adopt the approach of~\cite{Seibold17_IWDW} to manage the background. This method involves copying the face of the morphed image into the background of one aligned input image with a smooth transition for the low spatial frequency components of the image and a sharp cut for the high spatial frequency part of the image between these two images. Ghosting artifacts occur when structures in the input images are not properly aligned, e.g.~the nostrils have a different shape. Our approach can be seamlessly integrated into the morphing pipeline after the key-point-based alignment and before the additive blending.

\subsection{Problem Formulation and Optimization}
Pixel-wise alignment tasks are classically solved using the concept of the brightness constancy assumption~\cite{Horn1981}. Techniques based on this assumption aim to find a pixel warping from one image to another, minimizing the intensity difference between the warped and the target image.
Likewise, we are looking for a warping that maps similar structures to the exactly same shape and same location, but focuses on characteristic structures to estimate the warping, e.g.~borders of facial features such as specular highlights or the iris, instead of operating on intensity differences. Directly minimizing the intensity differences would lead to even worse aligned faces due to different skin tones or brightness variation. Thus, we first apply a spatially high-pass filter and only retain the high frequency information for our warping calculation. 

We calculate two independent warping functions to warp each image $I_1$ and $I_2$  independently to intermediate aligned images. 
With $I(\mathbf{p})$ being the pixel intensity of an image at a pixel position $\mathbf{p} \in \mathcal{N}^2$, the warped image $\tilde I$ can be defined as
\begin{equation}
 \tilde I(\mathbf{p}) = I(\mathbf{p} + w(\mathbf{p};\theta)),
\end{equation}
with $\theta$ being the warp parameters, i.e. the $x-$ and $y-$ offsets per pixel.

The loss function for the data term of the alignment is
\begin{dmath}
	\mathcal{L}_{d} = (\theta_1, \theta_2) \sum_{\mathbf{p} \in \mathcal{P}} \left | I_1(\mathbf{p} + w(\mathbf{p}; \theta_1)) - I_2(\mathbf{p} + w(\mathbf{p}; \theta_2))\right |_2^2,
\end{dmath}
 with $\mathcal{P}$ being the set of all pixel positions in the images. 
As this is an ill-posed problem, we add additional regularization terms that penalize the offset difference of neighboring pixels.
 \begin{equation}
 	\mathcal{L}_{s}(\theta) = \sum_{(\mathbf{p_1}, \mathbf{p_2}) \in \mathcal{P}_n} \left| w(\mathbf{p_1};\theta) - w(\mathbf{p_2};\theta) \right|_2^2
 	\label{eq:OFSmooth}	
 \end{equation}
 with $\mathcal{P}_n$ being neighboring pixels pairs, such that the second pixel is right or below the first pixel.
 \begin{equation}
 	\mathcal{L}_{b}(\theta) = \sum_{\mathbf{p} \in \mathcal{P}_b} \left| w(\mathbf{p};\theta) \right|_2^2
 	\label{eq:OFBorder}	
 \end{equation}
 with $\mathcal{P}_n$ being the pixels at the border of the image or region of interest that is optimized.
 
 The cost function to be minimized can thus be written as
 \begin{dmath}
 	\mathcal{L}(\theta_1, \theta_2) = \mathcal{L}_{d}(\theta_1, \theta_2) + \lambda \mathcal{L}_{s}(\theta_1)  + \lambda \mathcal{L}_{s}(\theta_2) + \lambda \mathcal{L}_{b}(\theta_1)  + \lambda \mathcal{L}_{b}(\theta_2),
 	\label{eq:OFCostFcn}
 \end{dmath}
 with $\lambda$ being a weighting factor for the smoothness term. 
 
 We minimize equation (\ref{eq:OFCostFcn}) using a Gauß-Newton algorithm. 
 Minimizing equation (\ref{eq:OFCostFcn}) is a non-linear optimization problem, since the data term, in particular, $I_1(\mathbf{p})$ and $I_2(\mathbf{p})$ are usually non-linear. However, since the images are already pre-aligned, a large warp is not expected and $I_1(\mathbf{p})$ and $I_2(\mathbf{p})$ are assumed to behave partly linear for small changes. During each iteration we thus minimize the following system
\begin{align}
	&\min_{w_{1,x}, w_{1,y}, w_{1,x}, w_{2,y}} \left\lVert
	\mathbf{A}
 \cdot 
        \mathbf{w}
 - \begin{bmatrix}
		\mathbf{i_2} - 	\mathbf{i_1}\\
		\mathbf{0}
	\end{bmatrix}
	\right\rVert,\\
 &\text{with } A = \begin{bmatrix}
			G_{1,x} & G_{1,y} & -G_{2,x} & -G_{2,y}\\
            \mathbf{P} \\
            &\mathbf{P} \\
            &&\mathbf{P} \\
            &&&\mathbf{P}
   \end{bmatrix}\\
   &\text{and }\mathbf{w}=\begin{bmatrix} \mathbf{w}_{1,x}^T & \mathbf{w}_{1,y}^T & \mathbf{w}_{2,x}^T & \mathbf{w}_{2,y}^T
   \end{bmatrix}^{T}
   \label{eq:SLE}
\end{align}
and $\mathbf{i_n}$ being the vectorized images $I_n$, $G_{n,x}$/$G_{n,x}$ diagonal matrices that contain the image gradient of $I_n$ in x-/y direction, $\mathbf{w}_{n,x}/\mathbf{w}_{n,y}$ the pixel motion in x-/y-direction and $P$ a sparse matrix that describes the smoothness term 
as defined in Equations \eqref{eq:OFSmooth} and \eqref{eq:OFBorder} scaled by $\sqrt{\lambda}$. It contains for every unordered pair of neighboring pixel one sparse row with $\sqrt{\lambda}$ at the column that represents the left or upper pixel and $-\sqrt{\lambda}$ at the column that represents the other pixel. For every border pixel, there is one row with only a non-zero entry in the column that represents that pixel, with a value of $\sqrt{\lambda}$.
The optimal solution to this problem can be obtained by solving 
\begin{equation}
	A^TA \mathbf{w} = A^T  [\mathbf{i_2} - \mathbf{i_1} \ \mathbf{0}]^T.
	\label{eq:linOpt}
\end{equation}
The matrix $A^TA$ is sparse but very large. Instead of explicitly setting it up, we utilize the Minimal Residual (MINRES) method to numerically solve equation (\ref{eq:linOpt}) \cite{MINRES75}. The MINRES method tackles the minimization problem through an iterative approach, requiring only a procedure for right-multiplication of arbitrary vectors $\mathbf{x}$ with the matrices $A$ and $A^T$. If we treat the vector $\mathbf{x}$ as an image, the multiplications related to the data term can be performed through pixel-wise operations with the image gradients. Similarly, the smoothness term can be implemented using convolution techniques.

\subsection{Examples of Improved Morphed Face Images}
\begin{figure}
    \centering
    \begin{subfigure}[b]{0.45\textwidth}
    \begin{tikzpicture}[spy using outlines={rectangle}, node distance=2mm and 0mm, inner sep=0mm]
		\node (I) {\includegraphics[width=0.45\textwidth]{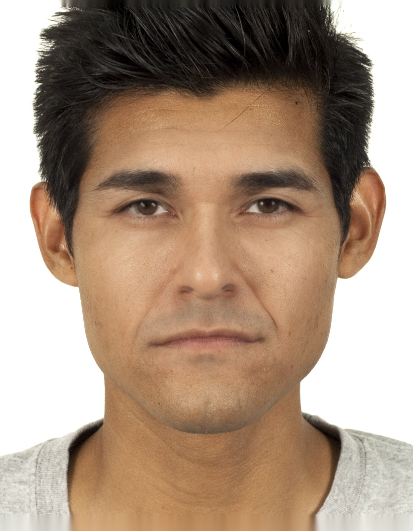}};
        \spy[red,magnification=4,size=0.3\textwidth]  on (0.0, -0.15) in node [below= of I](spy1);
    \end{tikzpicture}		
     \begin{tikzpicture}[spy using outlines={rectangle}, node distance=2mm and 0mm, inner sep=0mm]   
      \node(I2) {\includegraphics[width=0.45\textwidth]{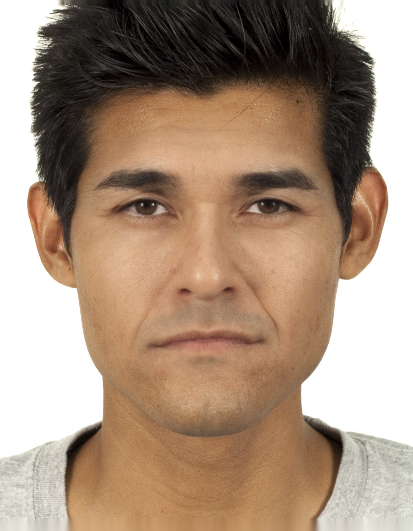}};
        \spy[red,magnification=4,size=0.3\textwidth]  on (0.0, -0.15) in node [below= of I2](spy2);
	\end{tikzpicture}		
    \label{fig:improvementsExamplesNose}
 \end{subfigure}\\[2ex]
  \begin{subfigure}[b]{0.45\textwidth}
    \begin{tikzpicture}[spy using outlines={rectangle}, node distance=2mm and 0mm, inner sep=0mm]
		\node (I) {\includegraphics[width=0.45\textwidth]{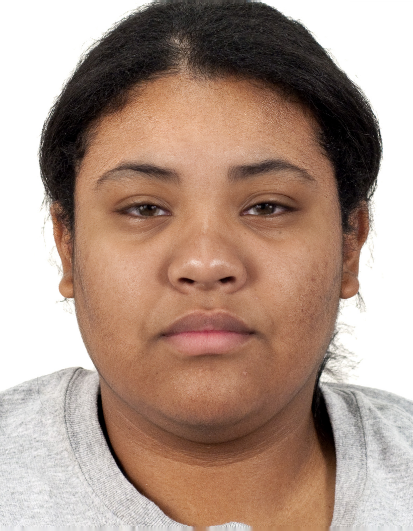}};
        \spy[red,magnification=5,size=0.3\textwidth]  on (-0.5, 0.5) in node [below= of I](spy1);
    \end{tikzpicture}		
     \begin{tikzpicture}[spy using outlines={rectangle}, node distance=2mm and 0mm, inner sep=0mm]   
      \node(I2) {\includegraphics[width=0.45\textwidth]{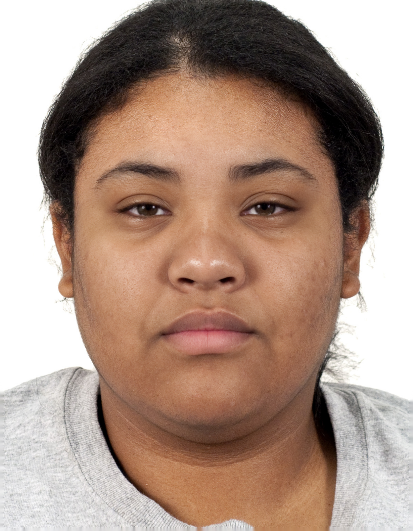}};
        \spy[red,magnification=5,size=0.3\textwidth]  on (-0.5, 0.5) in node [below= of I2](spy2);
	\end{tikzpicture}
    \label{fig:improvementsExamplesEye}
 \end{subfigure}
    \caption{Examples of different ghosting artifacts for a simple morphed face image (left) and our improved approach (right). The artifacts are avoided by the proposed alignment method.}
    \label{fig:improvementsExamples}
\end{figure}

Figure \ref{fig:improvementsExamples} shows further examples of our proposed method. The first example demonstrates the effectiveness of our method to avoid  a morphing artifact around the nostrils. This particular ghosting artifact commonly occurs with automatic face morphing pipelines, since the upper part of the nostrils is not estimated by standard facial landmark detector such as \cite{Kazemi2014}, which is often used due to its availability in DLib~\cite{king09}. 
Another often arising ghosting artifact is caused by misaligned specular highlights in the eyes as shown the second example. Again, these artifacts are avoided by the proposed method. 
\section{EXPERIMENTAL SETUP}
\subsection{Datasets}
For the training and evaluation of detectors, we compiled a large dataset of bona fide images from various sources and generated morphed face images using different methods, as described below.

We collected images from publicly available datasets, including BU-4DFE~\cite{BU4DFE}, CFD~\cite{CFD}, CFD-India~\cite{CFDINDIA}, CFD-MR~\cite{CFDMR}, FERET~\cite{FERET}, MR2~\cite{MR2}, FRLL~\cite{DeBruine2021}, PUT~\cite{PUT}, scFrontal~\cite{scFace}, SiblingDB~\cite{Vieira2014}, Utrecht\footnote{https://pics.stir.ac.uk/ \label{xx}}, YAWF~\cite{DeBruine2017}, RADIATE~\cite{Conley18}, Ecua~\cite{Aviles2019}, CUFS~\cite{Wang09}, Iranian Women\textsuperscript{\ref{xx}}, AMFD~\cite{Chen21}, stir\textsuperscript{\ref{xx}}, FED~\cite{Aifanti19} and FRGCv2~\cite{FRGC}. Additionally, we used in-house datasets and acquired additional face images through search engines. All images underwent manual checks to ensure that the subjects were in a neutral pose, looking directly into the camera, free from occlusions, and that a minimum inter-eye distance of 90 pixels was maintained. Furthermore, each subject was included only once in the dataset.

The FRGCv2 and FRLL datasets were exclusively selected for testing purposes. The other datasets, referred to as the mixed dataset, were divided into a training set with 70\% of all images and a testing and validation set with 15\% each. The mixed dataset consists of about 9,200 images with 6,400 images used for training, 1,440 for testing, and 1,400 for validation. The FRLL has 102 bona fide image with a neutral pose. From the FRGCv2 set, we utilized about 1,441 uniformly illuminated images with a neutral head pose and a uniform background for morph generation and further 1,726 images as reference images for the evaluation of the attack success on facial recognition systems.

Before using the images for training and evaluation, the faces were cropped such that they show the head and parts of the shoulder, as recommended by the ICAO~\cite{ICAO18PortraitQuality} for facial images stored on passports. After cropping, the images were resized to 513x431, which is a common size for passports~\cite{Neubert18}.  

\subsection{Morphed Face Images Generation}
We created morphed face images using five different methods or combinations of methods.
The simple morphs were generated using the pipeline from \cite{Seibold20}. The \textit{ST} morphs are an improved version of the simple morphs, incorporating  the style-transfer-based improvement described in \cite{Seibold19}. We refer to the simple morphs improved with our proposed pixel-wise alignment method as \textit{PW} morphs. When using both of the methods we refer to them as \textit{PWST} morphs. For creating GAN-based morphs, we use the method of \cite{Zhang21} and refer to them as \textit{MIP2} morphs. To generate the \textit{MIP2} morphs, we used the implementation of \cite{Sarkar22}.

To select suitable pairs for generating morphed face images, we followed the protocol in \cite{Scherhag20} for the FRGCv2 dataset and in \cite{Neubert18} for the FRLL dataset. For the generation of morphed face images from the other dataset, we selected the pair such that they are both from the same dataset and their gender and ethnicity match. The number of morphs and bona fide images in the respective set is the same. During testing and validation, the data is augmented by horizontal flipping. The validation set is specifically used for the evaluation of the epochs of detectors based on Deep Neural Networks (DNNs), selecting the best performing model.

\subsection{Detectors}
We investigate the impact of our proposed ghosting artifact prevention method on the detection rates of five detectors. One of the detectors is based on an ensemble of features and utilizes a probabilistic CRC for classification~\cite{Raghavendra19}. The remaining detectors employ DNNs~\cite{Seibold21}. All methods only operate on the inner part of the face as proposed by their authors.

The DNN-based detectors in our study use one output neuron and are all trained using a binary cross-entropy loss. The detector \textit{VGG-A naïve} uses the VGG-A architecture and \textit{Xception} the Xception architecture, which has demonstrated effectiveness in detecting Deep Fakes~\cite{Malolan20}. The \textit{Feature Focus}~\cite{Seibold21} detector incorporates an additional loss that activates half of the neurons in the last convolutional layer strongly for morphed face images, and the other half for bona fide images.
Inspired by the work of \cite{Raghavendra19} on effects of color spaces on the performance of morphing detectors, we tested the most promising detector on images in the HSV color space. These are denoted with (HSV). The Feature Focus detector in HSV color space, trained only randomly compressed images, was also submitted to the Face Morphing Detection challenge of the NIST. It showed an outstanding performance and took first place in different categories~\cite{FATE}.

The \textit{Feature Ensemble} detector \cite{Venkatesh20} splits the images into two different color spaces and calculates a Laplacian pyramid with three levels. For each of the resulting images, a Histogram of Gradients, Binarized Statistical Image Features and Local Binary Pattern are calculated and a probabilistic CRC classifier is employed.

\section{RESULTS}
In the following, each subsection will answer one of the research questions presented in the Introduction.

\begin{table*}[!htb]
\centering
	\caption{EER and BPCER@APCER=5\% for training on simple morphs only
 to analyze the effect of different morphed face image improvement methods and a GAN-based generation method. The best performing attack method is highlight in bold and the second best are underlined. The PWST morphing method, which is our proposed method in combination with an improvement based on style transfer~\cite{Seibold19}, achieves in all cases the highest or second highest error rates and thus the improved morphed face images are harder to detect than the simple morphed face images without any improvement applied.}
			\begin{tabular}{|l|ccccc|ccccc|}
                \hline
				\multicolumn{1}{|r|}{Morph}& \multicolumn{5}{|c|}{Equal-Error-Rates[\%]} & \multicolumn{5}{c|}{BPCER[\%]@APCER=5\%} \\
		          \multicolumn{1}{|r|}{Method}&simple & PW & ST & PWST & MIP2 & simple & PW & ST & PWST & MIP2 \\		
                Dataset& & (ours) & & (ours+ST) & & & (ours) & & (ours+ST) & \\
                \hline
                &\multicolumn{10}{|c|}{Detector: \textit{Feature Ensemble}~\cite{Venkatesh20}}\\
                Mixed Set &5.03 & 5.73 & \underline{20.67} & \textbf{21.68} & 15.24 & 5.24 & 6.56 & \underline{49.51} & \textbf{49.83} & 37.72\\
				  FRLL      &1.96 & 1.96 & 10.29 & \underline{11.27} & \textbf{12.75} & 0.00 & 0.49 & 15.69 & \underline{17.16} & \textbf{25.00} \\
				FRGCv2      &3.46 & 4.93 & 14.13 & \underline{15.96} & \textbf{22.36} & 2.03 & 4.78 & 29.67 & \underline{36.69} & \textbf{60.92}\\
				\hline
                &\multicolumn{10}{|c|}{Detector: \textit{VGG-A}~\cite{Simonyan15}}\\
                Mixed Set &0.59 & 1.67 & 11.61 & \underline{16.50} & \textbf{22.22} & 0.14 & 0.73 & 21.01 & \underline{30.87} & \textbf{78.33}\\
				  FRLL      &0.00 & 0.00 & 0.49 & \underline{1.96} & \textbf{13.24} & 0.00 & 0.00 & 0.00 & \underline{0.49} & \textbf{61.76}\\
				FRGCv2      &0.21 & 1.17 & 2.10 & \underline{6.46} & \textbf{9.45} & 0.00 & 0.25 & 1.37 & \underline{7.62} & \textbf{17.99}\\
				\hline
                &\multicolumn{10}{|c|}{Detector: \textit{Xception}~\cite{Chollet17}}\\
                Mixed Set  &0.38 & 1.39 & 6.78 & \underline{10.70} & \textbf{11.01} & 0.07 & 0.24 & 8.65 & \underline{18.30} & \textbf{30.10}\\
				  FRLL       &0.05 & 0.71 & 3.43 & \underline{7.84} & \textbf{12.75} & 0.00 & 0.00 & 1.96 & \underline{12.25} & \textbf{43.63}\\
				FRGCv2       &0.47 & 2.34 & 6.41 & \textbf{12.45} & \underline{10.53} & 0.05 & 1.32 & 7.77 & \underline{23.07} & \textbf{24.24}\\
				\hline
                &\multicolumn{10}{|c|}{Detector: \textit{Feature Focus (RGB)}~\cite{Seibold21}}\\
                Mixed Set  &0.73 & 1.29 & 11.64 & \textbf{15.11} & \underline{13.40} & 0.07 & 0.63 & 21.63 & \underline{29.65} & \textbf{51.15}\\
				  FRLL       &0.00 & 0.00 & 0.49 & \underline{1.96} & \textbf{10.31} & 0.00 & 0.00 & 0.00 & \underline{1.96} & \textbf{29.90}\\
				FRGCv2       &0.21 & 0.46 & 1.84 & \underline{6.05} & \textbf{7.99} & 0.00 & 0.20 & 0.71 & \underline{7.27} & \textbf{20.58}\\
				\hline
                &\multicolumn{10}{|c|}{Detector: \textit{Feature Focus (HSV)}~\cite{Seibold21}}\\
                Mixed Set  &0.97 & 1.04 & \underline{12.23} & \textbf{13.45} & 7.15 & 0.07 & 0.14 & \underline{21.84} & \textbf{26.88} & 12.33\\
				  FRLL       &0.00 & 0.00 & 1.03 & \underline{1.56} & \textbf{2.45} & 0.00 & 0.00 & 0.00 & 0.00 & \textbf{0.49}\\
				FRGCv2       &0.00 & 0.00 & 4.67 & \textbf{6.96} & \underline{5.55} & 0.00 & 0.00 & 4.62 & \textbf{9.71} & \underline{7.06}\\
				\hline
        \multicolumn{11}{l}{simple: \cite{Seibold20}; ST: \cite{Seibold19}; MIP2: \cite{Zhang21}}
        \end{tabular}
	\label{tab:ErrorRatesSimple}
\end{table*}
\subsection{Are Our Improved Morphs Harder to Detect?}

To examine the impact of the ghosting artifact removal and address \textbf{R1}, we trained all detectors on simple morphs and evaluated them on all types of morphs.
We report the detectors' performance using Attack Presentation Classification Error Rates (APCER) and Bona fide Presentation Classification Error Rates (BPCER) as defined in ISO/IEC 30107-3 \cite{PAISO} and Equal Error Rates (EER). The BPCER is reported at a fixed APCER of 5\%.
Table \ref{tab:ErrorRatesSimple} reveals that the removal of ghosting artifacts has only a small impact on the detection, in contrast to the \textit{ST} improvement or the utilization of GANs for morph generation. However, these morphs are still harder or at least as hard to detect. While the difference in the EERs for the simple morphs and the \textit{PW} morphs is always smaller than 2\%, the EERs for the \textit{ST} morphs are more than 10\% larger and the EER for the \textit{MIP2} morphs is even up to 20\% larger for the \textit{VGG-A naïve} detector.
In combination with the style-transfer-based improvement, however, the error rates notably increase compared to using style-transfer only for the improvement. 

\subsection{Can the Detectors Adapt to the New Challenge?}
To assess whether the detectors can adapt to the proposed improved morphs and the other types of morphs (\textbf{R2}), we added \textit{PWST} and \textit{MIP2} morphs to the training data. The results are shown in Table \ref{tab:ErrorRatesMixed}. For the DNN-based detectors, the error rates significantly decreased for the \textit{PWST} and \textit{MIP2} morphs in nearly all cases. The error rates for the PW and ST morphs drop in most cases, but the rates for the simple morphs slightly increase in most cases. The \textit{Feature Ensemble} detector shows the largest error rates and has also the strongest increase in error rates for the simple morphs. To answer \textbf{R2}: The DNN-based detectors can easily adapt to the improved and to the \textit{MIP2} morphs by just adding examples of these morphs to the training data. The \textit{Feature Focus (HSV)} detector shows the best performance. 
\begin{table*}[!hbt]
\centering
	 \caption{EER and BPCER@APCER=5\% for training on simple, \textit{PWST} and \textit{MIP2} morphs to analyze if the detectors can adapt to the threat of improved and GAN-based morphed face images. The best-performing attack method is highlight in bold and the second best are underlined. The DNN-based detectors seem to be able to learn other traces of forgery to distinguish between bona fide and morphed face images.}
  
		\begin{tabular}{|l|ccccc|ccccc|}
                \hline
				\multicolumn{1}{|r|}{Morph}& \multicolumn{5}{|c|}{Equal-Error-Rates[\%]} & \multicolumn{5}{c|}{BPCER[\%]@APCER=5\%} \\
		          \multicolumn{1}{|r|}{Method}&simple & PW & ST & PWST & MIP2 & simple & PW & ST & PWST & MIP2 \\		
                Dataset& & (ours) & & (ours+ST) & & & (ours) & & (ours+ST) & \\
                \hline
                &\multicolumn{10}{|c|}{Detector: \textit{Feature Ensemble}~\cite{Venkatesh20}}\\
                Mixed Set &14.21 & 13.97 & \underline{14.52} & \textbf{14.87} & 4.34 & \textbf{29.27} & \underline{28.72} & 28.30 & \underline{28.72} & 4.06\\
				  FRLL      &\textbf{7.84} & \textbf{7.84} & 5.88 & 5.98 & 0.53 & \textbf{10.78} & \textbf{10.78} & 6.86 & 6.86 & 0.00\\
				FRGCv2      &10.06 & 11.00 & \underline{13.72} & \textbf{14.48} & 2.59 & 17.89 & 22.21 & \underline{28.81} & \textbf{32.72} & 1.27\\
				\hline
                &\multicolumn{10}{|c|}{Detector: \textit{VGG-A}~\cite{Simonyan15}}\\
                Mixed Set &1.15 & 0.69 & \textbf{2.19} & \underline{1.74} & 0.87 & 0.28 & 0.17 & \textbf{1.01} & \underline{0.59} & 0.14\\
				  FRLL      &0.00 & 0.00 & 0.00 & 0.00 & 0.00 & 0.00 & 0.00 & 0.00 & 0.00 & 0.00\\
				FRGCv2      &1.95 & 2.08 & \textbf{5.71} & \underline{5.29} & 1.14 & 0.56 & 0.81 & \textbf{6.30} & \underline{5.44} & 0.15\\
				\hline
                &\multicolumn{10}{|c|}{Detector: \textit{Xception}~\cite{Chollet17}}\\
                Mixed Set  &0.56 & 0.31 & \textbf{1.25} & \underline{0.90} & 0.52 & 0.03 & 0.00 & \textbf{0.21} & \underline{0.14} & 0.00\\
				  FRLL       &0.00 & 0.00 & \underline{0.07} & 0.02 & \textbf{0.49} & 0.00 & 0.00 & 0.00 & 0.00 & 0.00\\
				FRGCv2       &1.37 & 1.84 & \underline{5.13} & \textbf{5.24} & 0.41 & 0.41 & 0.66 & \underline{5.28} & \textbf{5.64} & 0.00 \\
				\hline
                &\multicolumn{10}{|c|}{Detector: \textit{Feature Focus (RGB)}~\cite{Seibold21}}\\
                Mixed Set  &1.20 & 1.84 & \underline{3.50} & \textbf{4.61} & 0.83 & 0.28 & 0.92 & \underline{2.44} & \textbf{4.05} & 0.14\\
				  FRLL       &0.00 & 0.00 & 0.00 & 0.00 & 0.00 & 0.00 & 0.00 & 0.00 & 0.00 & 0.00\\
				FRGCv2       &1.69 & 2.70 & \underline{5.39} & \textbf{6.77} & 0.83 & 0.71 & 1.88 & \underline{5.89} & \textbf{8.59} & 0.15\\
				\hline
                &\multicolumn{10}{|c|}{Detector: \textit{Feature Focus (HSV)}~\cite{Seibold21}}\\
                Mixed Set  &1.04 & 0.73 & \underline{1.15} & \textbf{1.18} & 0.49 & 0.28 & 0.21 & \underline{0.31} & \textbf{0.38} & 0.07\\
				  FRLL       &0.00 & 0.00 & 0.00 & 0.00 & 0.00 & 0.00 & 0.00 & 0.00 & 0.00 & 0.00\\
				FRGCv2       &0.00 & 0.00 & \textbf{0.15} & \underline{0.10} & 0.00 & 0.00 & 0.00 & 0.00 & 0.00 & 0.00\\
				\hline
        \multicolumn{11}{l}{simple: \cite{Seibold20}; ST: \cite{Seibold19}; MIP2: \cite{Zhang21}}
        \end{tabular}
  \label{tab:ErrorRatesMixed}
\end{table*}

\begin{table*}
\centering
		\caption{MMPMR@FAR0.1\% for FRGCv2 dataset to analyse if the morphed face images portray two different subjects. The improvement methods seem to have only a marginal impact on the biometric properties of the face images and after applying the improvement methods (ours + ST), the attacks are more successful than the baseline (simple). Whether the biometric quality of the GAN-based morphs (MIP2) is better or much worse than these of the keypoint-based morphs strongly depends on the facial recognition system used for the evaluation.}
		\setlength{\tabcolsep}{4pt}
		\begin{tabular}{|l|cc|ccc|}
			\hline
            & \multicolumn{2}{|c|}{Bona Fide} & \multicolumn{3}{|c|}{MMPMR@FAR0.1\%} \\ 
			&FRR & FAR & simple & PWST & MIP2 \\
            & & & \cite{Seibold20} & (ours + ST) & \cite{Zhang21}  \\
			\hline
			ArcFace & 1.18\% & 0.1\% & 31.48\% & 32.57\% & \textbf{37.97}\%\\
			COTS    & 0.00\% & 0.02\% & 48.50\% & \textbf{48.65}\% & 17.63\%\\
			\hline
		\end{tabular}
	\label{tab:RMMAR}
\end{table*}

\subsection{Do the Improved Morphs Still Impersonate Two Different Subjects?}
To address \textbf{R3}, we evaluated the biometric quality using the MinMax-Mated Morph Presentation Match Rate (MMPMR)~\cite{Scherhag17} on the FRGCv2 dataset using the protocol of \cite{Scherhag20}. Two different facial recognition systems were employed for the evalution: 
An implementation of ArcFace\footnote{https://github.com/mobilesec/arcface-tensorflowlite} and a commercial off-the-shelf (COTS) system. The false acceptance rate (FAR) threshold for the COTS system was determined based on its manual and for the ArcFace system calculated from the FRGCv2 dataset. Table \ref{tab:RMMAR} shows that the improvement methods have only a minor effect on the biometric quality and they even improve the success rate of the attacks slightly. Another interesting finding is that the \textit{MIP2} morphs are much better at tricking the ArcFace system than the COTS system and that they perform much worse on the COTS than the other morphs do. 

\subsection{Do the Improvements Make a Difference for JPG-Compressed Images?}
\begin{table*}[!hbt]
\centering
\caption{Error rates of the Feature Focus (HSV) detector for compressed images. 
The best performing attack method is highlight in bold and the second best are underlined. Also for the JPG-compressed images, the \textit{PWST} and MIP2 morphs are much harder to detect than the simple morphs. The detector can adapt to the improved morphs, but does still perform much worse than on the simple attacks.}
		\begin{tabular}{|l|ccccc|ccccc|}
                \hline
				\multicolumn{1}{|r|}{Morph}& \multicolumn{5}{|c|}{Equal-Error-Rates[\%]} & \multicolumn{5}{c|}{BPCER[\%]@APCER=5\%} \\
		          \multicolumn{1}{|r|}{Method}&simple & PW & ST & PWST & MIP2 & simple & PW & ST & PWST & MIP2 \\		
                Dataset& & (ours) & & (ours+ST) & & & (ours) & & (ours+ST) & \\
                \hline
                &\multicolumn{10}{|c|}{Detector: \textit{Festure Focus (HSV)}~\cite{Seibold21} train on simple only}\\
                Mixed Set &3.41 & 6.60 & 17.93 & \underline{24.67} & \textbf{25.14} & 2.12 & 8.54 & 37.67 & \underline{55.17} & \textbf{76.63}\\
				  FRLL      &2.60 & 5.59 & 14.22 & \underline{21.57} & \textbf{32.36} & 0.00 & 6.37 & 22.55 & \underline{48.53} & \textbf{79.41}\\
				FRGCv2      &1.02 & 3.76 & 11.85 & \underline{20.38} & \textbf{21.80} &0.15 & 3.00 & 22.26 & \underline{46.19} & \textbf{78.81}\\
				\hline
                \multicolumn{1}{|r|}{Method}&simple & PW & ST & PWST & MIP2 & simple & PW & ST & PWST & MIP2 \\		
                Dataset& & (ours) & & (ours+ST) & & & (ours) & & (ours+ST) & \\
                \hline
                &\multicolumn{10}{|c|}{Detector: \textit{Festure Focus (HSV)}~\cite{Seibold21} train on simple, PWST and MIP2}\\
                Mixed Set &4.83 & 5.11 & \underline{10.46} & \textbf{11.19} & 6.46 & 4.65 & 5.21 & 1\underline{8.61} & \textbf{18.72} & 7.74\\
				  FRLL      &3.93 & 6.37 & \underline{10.78} & \textbf{14.78} & 11.42 & 3.43 & 7.84 & \underline{17.65} & \textbf{26.47} & 21.57\\
				FRGCv2      &2.80 & 3.30 & \textbf{7.67} & \underline{7.62} & 7.37 &1.22 & 1.73 & \underline{10.82} & \textbf{11.48} & 9.96\\
				\hline
    \multicolumn{11}{l}{simple: \cite{Seibold20}; ST: \cite{Seibold19}; MIP2: \cite{Zhang21}}
    \end{tabular}
	\label{tab:EERBPCERMixedCompressed}
\end{table*}
Table \ref{tab:EERBPCERMixedCompressed} shows the error rates of the Feature Focus (HSV) detector trained and tested on compressed images to answer \textbf{R4}. We used a compression rate that targets a file size between 15kB and 20kB, which is the typically mandatory and reserved size for storing facial image on the passport chips~\cite{ICAO15Size}. The error rates for the morphs improved by our ghosting artifact prevention method are larger than those for the simple morphs. In combination with the style-transfer improvement, the error rates are even larger compared to when only one of them is used. Thus, our proposed ghosting artifact prevention also affects the detection rates in compression face images.

\section{SUMMARY AND DISCUSSION}
In this paper, we introduced a ghosting artifact prevention method that can be integrated into key-point based face morphing pipelines. The prevention of ghosting artifacts can be performed manually by an attacker, but this is not feasible for large training or evaluation datasets. Our approach effectively prevents ghosting artifacts without compromising the biometric quality of the morphs. Furthermore, it poses a greater challenge for MAD techniques to detect these improved morphs. In combination with the style-transfer-based improvement method of \cite{Seibold19}, the resulting morphed face images provide a new challenge for MAD techniques. One of its biggest advantages compared to GAN-based methods is that our method can produce morphs in any resolution, while the resolution of GAN-morphs is limited by the GAN's architecture. 
Furthermore, the keypoints-based approach allows a better control over the morphed face image generation process. This control includes factors such as balancing the influence of each individual input image on the resulting morphed face image (blending factor), specifying which regions should be blended, and other parameters. The keypoint-based morphing pipeline closely aligns with the approach an attacker might use to create a morphed face images, utilizing publicly-available tools for warping and blending.
By incorporating the improved morphs into the training data, we observe enhanced detection performance for these improved morphs. This further highlights the effectiveness and practical significance of our approach in improving the detection capabilities of MAD techniques. In future work, we plan to study the impact of the our proposed improvement on detectors that analyze the shape of reflections, given that pixel-wise alignment changes the face's geometry~\cite{Seibold18_EUSIPCO}.

\section*{\uppercase{Acknowledgements}}
This work has recieved partial funding by the German Federal Ministry of Education and Research (BMBF) through the Research Program FAKEID under Contract no. 13N15735, as well as the Fraunhofer Society in the Max Planck-Fraunhofer collaboration project NeuroHum.

\bibliographystyle{apalike}
{\small
\bibliography{lniguide}}

\end{document}